\pdfoutput=1
\documentclass[11pt]{article}

\usepackage{acl}

\usepackage{times}
\usepackage{latexsym}

\usepackage[T1]{fontenc}
\usepackage[utf8]{inputenc}

\usepackage{microtype}

\title{Structural Transfer Learning in NL-to-Bash Semantic Parsers}

\author{Kyle Duffy \\
  University of Oxford \\
  \texttt{kylejduffy3@gmail.com} \\ \And
  Satwik Bhattamishra \\
  University of Oxford \\
  \texttt{satwik.bmishra@stx.ox.ac.uk} \\ \AND
  Phil Blunsom \\
  University of Oxford \\
  \texttt{phil.blunsom@cs.ox.ac.uk} \\}

\usepackage{amsthm, amssymb, amsmath, bbm}

\usepackage{enumitem}
\newlist{checklist}{itemize}{2}
\setlist[checklist]{label=$\square$}
\usepackage{pifont}

\newenvironment{itemize*}%
	{\begin{itemize}%
		\setlength{\itemsep}{0pt}%
		\setlength{\parskip}{0pt}%
		\setlength{\parsep}{0pt}}%
	{\end{itemize}}
\newenvironment{enumerate*}%
	{\begin{enumerate}%
		\setlength{\itemsep}{0pt}%
		\setlength{\parskip}{0pt}%
		\setlength{\parsep}{0pt}}%
	{\end{enumerate}}
\newenvironment{checklist*}%
	{\begin{checklist}%
		\setlength{\itemsep}{0pt}%
		\setlength{\parskip}{0pt}%
		\setlength{\parsep}{0pt}}%
	{\end{checklist}}
\DeclareMathOperator{\ppl}{ppl}

\usepackage{lipsum}
\usepackage{booktabs}

\begin{document}
\maketitle

\section{Introduction}

Semantic parsing is the task of extracting semantic meaning from natural language. The meaning is parsed into a machine-readable format, which is typically a programming language. In this paper, we consider a semantic parsing system for the natural language to Bash (NLBash) task.

Modifying pre-training data with synthesized samples or external information can improve performance of semantic parsers; synthesizing pre-training data with synchronous context-free grammars has improved performance on the natural language to SQL (NLSQL) task \citep{yu2021grappa,yu2021score,gap}, and incorporating context from external web data has proven effective for the natural language to Python (NLPython) task \citep{xu-etal-2020-incorporating}.

However these systems are often language-specific or rule-based, and end up being difficult to scale, imprecisely targeted, and monolingual.
A deeper understanding of structural transfer to semantic parsing could suggest which datasets should be included during pre-training and broaden the scope of pre-training data to include multilingual or out-of-task samples.

We measure transferability to NLBash from 6 upstream tasks by computing a heuristic for semantic accuracy of Bash programs introduced by \citet{nlc2cmd}, and also accuracy and perplexity. We find that NLSQL transfers strongly to NLBash, NLPython has little marginal improvement over lexical mapping, and more data with longer upstream training does not always lead to more semantically accurate programs.

\section{Method}\label{sec:method}

We measure how well a transformer can model NLBash using weights learned from a different task. To do so, we follow \citet{papadimitriou-jurafsky-2020-learning} and freeze transformer weights after pre-training and then fine-tune a linear embeddings layer on NLBash.

Formally, an upstream source vocabulary $S_U$, upstream target vocabulary $T_U$, downstream source vocabulary $S_D$, and downstream target vocabulary $T_D$ are given along with their corresponding sets of sentences, which we denote as $S_U^*$, $T_U^*$, $S_D^*$, $T_D^*$.

A transformer model $Y$ of a fixed depth and hidden dimension is chosen and initialized. An upstream task $U$ and a downstream task $D$ are selected and given as a set of sentence pairings, $U\subseteq S_U^*\times T_U^*$, $D\subseteq S_D^*\times T_D^*$. The downstream task is split into a training set $D_{\text{train}}$, and a test set $D_\text{test}$.

An embedding layer $e_U = (e_U^s, e_U^t)$ is prepended to the transformer model $Y$, and the model is trained end-to-end on the upstream task to convergence. This is called the \emph{pre-training phase}.

The upstream embeddings $e_U$ are removed and a new embedding layer, $e_D = (e_D^s, e_D^t)$ is initialized and prepended to $Y$. The weights of $Y$ are frozen, but $e_D$ is trained on $D_\text{train}$ in order to fit a lexical alignment to the representations learned during pre-training. This phase is called the \emph{fine-tuning phase}. During this phase, only a linear embedding layer is learned so any higher order features must come from structural representations learned upstream. Freezing $Y$ prevents leakage from the downstream task that would dilute the structural content learned during pre-training.

Finally, the composite model $Y\circ e_D$ is paired with a (fixed) decoding algorithm and evaluated on $D_\text{test}$. This is called the \emph{evaluation phase}.

\subsection{Evaluation}

We evaluate three performance metrics: (1) an approximation of semantic accuracy from \citet{nlc2cmd}; this is a score between --100 (worst) and 100 (best) which we call the Bash Similarity Heuristic (BaSH); (2) accuracy, which is the proportion of words which are the model's maximum likelihood estimate; (3) perplexity ($\ppl$) computed on downstream target programs.

\subsection{Datasets}

\paragraph{Copy task.}
$\sim$10k synthetic source sequences paired with their image under a vocabulary permutation.

\paragraph{End-to-End.}
NLBash \citep{nlc2cmd}. This is equivalent to end-to-end training.

\paragraph{English to Python.}
The CoNaLa bitext corpus from \citet{conala}.

\paragraph{English to SQL.}
The Spider dataset \citep{spider}. We mask constants and omit schema context to reduce complexity without disturbing syntax.

\paragraph{SQL to English (SQLNL).}
NLSQL with sources and targets interchanged.

\paragraph{English to German.}
The WMT14 English to German translation task (En-De) as curated by \citet{luong-etal-2015-effective}.

\paragraph{Reversal.}
$\sim$10k synthetic source sequences paired with targets constructed by reversing word order in the source.

\section{Results and Discussion}

\begin{table}[h]
\centering
\begin{tabular}{lrrr}
\toprule
\textbf{Experiment} & \textbf{ppl} & \textbf{Acc.} & \textbf{BaSH}\\ \hline
Uniform Init. & 3.29 & 67.9 & --8.3  \\
Xavier Init. & 3.11 & 75.0 & 15.1 \\
End-to-End & 3.80 & 78.1 & 16.5 \\ \hline 
NLPython & 3.39 & 74.2 & 10.1 \\
NLSQL & 3.14 & 75.6 & {\bf 14.1} \\
SQLNL & 3.27 & 75.2 & 13.5 \\
En-De & {\bf 2.77} & {\bf 76.3} & 12.9 \\
Reversal & 3.33 & 73.2 & 7.2 \\
Copy & 3.24 & 73.8 & 11.1 \\ \bottomrule
\end{tabular}
\caption{Results.}
\label{tab:results}
\end{table}

We find that representations from NLSQL transfer strongly to NLBash as is shown in Table \ref{tab:results}. This corroborates the intuition that these tasks are structurally similar and is expected since NLBash consists largely of \verb|find| commands which have a similar structure to \verb|SELECT| statements in SQL.

Even so, the model pre-trained on NLSQL did not score better than a model with random weights. This reinforces the somewhat surprising effectiveness of random models under the Xavier initialization scheme.

The copy task also scores well despite its simplicity. Indeed, we observe no substantial improvement of NLPython over the copy task, which suggests that \emph{although NLPython exhibits semantic patterns such as loops and control flow, it provides little transferable gain over direct lexical mapping}.

Finally, $\ppl$ and accuracy only seem loosely related to semantic accuracy measured by BaSH.
En-De outperformed all upstream tasks on $\ppl$ and accuracy, however, considerably more compute was expended during pre-training. This indicates a correlation between expended compute and perplexity, and the remaining measurements suggest this trend may not be shared by semantic accuracy. To test this idea, we varied dataset size and training length.

\subsection{Varying Compute}\label{sec:compute}

\begin{table}[h]
\centering
\begin{tabular}{lcccr}
\toprule
\textbf{Samples} & \textbf{Steps} & \textbf{ppl} & \textbf{Acc.} & \textbf{BaSH} \\ \hline
10k & 2500 & 3.39 & 75.0 & 13.0 \\
10k & 7500 & 3.27 & 75.0 & 11.8 \\
100k & 2500 & 3.13 & 76.0 & \textbf{14.2} \\
100k & 7500 & 3.16 & 75.8 & 13.7 \\
4.4M & 2500 & 2.95 & 76.0 & 12.6 \\
4.4M & 10000 & \textbf{2.77} & \textbf{76.3} & 12.9 \\
\bottomrule
\end{tabular}
\caption{English to German, several compute samples.}
\label{tab:decompute}
\end{table}

We randomly downsample En-De to 10K and 100K samples and train for 2500 and 7500 steps and display results in Table \ref{tab:decompute}.

There is a clear correlation between larger datasets, more training steps, and better performance measured by perplexity and accuracy. 
Interestingly, the same trend is not shared by BaSH. We instead find that \emph{transferability of semantic representations actually appears to degrade with upstream training past a point}.

\subsection{Future Work}

A natural extensions is to include (more) multilingual and multimodal upstream data.
Further, the extent that transferability matures into performance of fine-tuned models is untested. It would be valuable to investigate how the convergence rate of unfrozen, fine-tuned models varies with (1) upstream task and (2) compute expended during pre-training.

Finally, it would be valuable for practitioners to understand the strength of the correlation between transferability (measured here by BaSH evaluated on frozen transformers) and fine-tuned performance.

% ==============================================================================
%   End Content
% ==============================================================================

% Entries for the entire Anthology, followed by custom entries
\bibliography{anthology,custom}

\appendix

\section{Model}
\label{sec:model}

We use a simple transformer encoder decoder stack \citep{vaswani2017} with 6 layers and 8 attention heads, trained with cross-entropy loss. The hidden size is 512 and feedforward layers have dimension 2048. All models are regularized with dropout and label smoothing rates of 0.1. All models are trained on two NVIDIA GeForce GTX 1080 Ti GPUs with 11GB of random access memory and converge well within one day of training.

\end{document}